# Computer-aided position planning of miniplates to treat facial bone defects

Jan Egger[1,2‡*], Jürgen Wallner[3‡], Markus Gall[1], Xiaojun Chen[4], Katja Schwenzer-Zimmerer[3], Knut Reinbacher[3], Dieter Schmalstieg[1]

**1** Institute for Computer Graphics and Vision, Faculty of Computer Science and Biomedical Engineering, Graz University of Technology, Graz, Austria, **2** BioTechMed-Graz, Graz, Austria, **3** Department of Oral & Maxillofacial Surgery, Medical University of Graz, Graz, Styria, Austria, **4** Institute of Biomedical Manufacturing and Life Quality Engineering, School of Mechanical Engineering, Shanghai Jiao Tong University, Shanghai, China

‡ These authors are joint first authors on this work.
* egger@tugraz.at

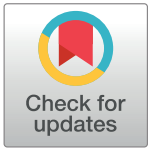

**Data Availability Statement:** All relevant data are uploaded to Figshare at URL: https://figshare.com/articles/Cranial_Defect_Datasets/4659565; DOI: https://doi.org/10.6084/m9.figshare.4659565.v1.

**Funding:** The work received funding from BioTechMed-Graz in Austria ("Hardware accelerated intelligent medical imaging") and the 6th Call of the Initial Funding Program from the Research & Technology House (F&T-Haus) at the Graz University of Technology (PI: DDr. Jan Egger). Dr. Xiaojun Chen receives support by the Natural

## Abstract

In this contribution, a software system for computer-aided position planning of miniplates to treat facial bone defects is proposed. The intra-operatively used bone plates have to be passively adapted on the underlying bone contours for adequate bone fragment stabilization. However, this procedure can lead to frequent intra-operatively performed material readjustments especially in complex surgical cases. Our approach is able to fit a selection of common implant models on the surgeon's desired position in a 3D computer model. This happens with respect to the surrounding anatomical structures, always including the possibility of adjusting both the direction and the position of the used osteosynthesis material. By using the proposed software, surgeons are able to pre-plan the out coming implant in its form and morphology with the aid of a computer-visualized model within a few minutes. Further, the resulting model can be stored in STL file format, the commonly used format for 3D printing. Using this technology, surgeons are able to print the virtual generated implant, or create an individually designed bending tool. This method leads to adapted osteosynthesis materials according to the surrounding anatomy and requires further a minimum amount of money and time.

## Introduction

The Reconstructions of facial deformations and defects due to bone fractures where two bone fragments are operatively stabilized, so called osteosynthesis, is part of a surgeon's daily life. More precisely, these bone fractures occur as a result of applied outer forces, like happening due to traffic accidents, results of tumor removal or deformation treatment [1]. One major cause for facial fractures in regions where winter sports are frequently practiced is skiing and snowboarding [2]. These two sports account for most of the facial injuries overall, since wearing a helmet is not mandatory thus resulting in a high percentage of practitioners not equipped with appropriate protection like helmets, even though high velocities are reached not able to Science Foundation of China (Grant No.: 81511130089) and the Foundation of Science and Technology Commission of Shanghai Municipality (Grants Nos.: 14441901002, 15510722200 and 16441908400).

**Competing interests:** The authors have declared that no competing interests exist.

be absorbed by the skull in case of collision. According to Gassner et al. [3], the average age of people transferred to the department of oral and maxillofacial surgery with facial injuries lies by 26 years with 50% between age 16 and 38. The group with the highest rate of injuries is the one of children between 7 and 12 years. Further, there is a high variation between the groups of men and woman suffering from soft tissue damages or bone fractures in the facial area due to skiing accidents, stating that men account for 65.3% of the cases whereas females show a number of 34.7% injured patients. Moreover, the types of injuries range from facial bone fractures to dento-alveolar traumas and soft tissue injuries. There is also a wide range of different mechanisms including simple falls, collisions with others or objects, a struck by equipment, lift accidents and various more. For a better understanding of these data, Table 1 gives a detailed overview of these findings.

However, accidents resulting in facial bone fractures require osteosynthesis where the application of varying osteosynthesis materials such as miniplates [4], [5] (Fig 1) is essential. As an example, Fig 2 shows an osteosynthesis of a mandibular angle fracture. In this case two adaptive plates where used, one aligned to the upper mandibular border (Linea Oblique) and the other on the lower border of the mandible. Additionally Fig 3 shows an x-ray image of a case suffering from such a mandibular angle fracture. To the left, the defect is shown before treatment. On the right image the same patient after bone repositioning and miniplate application is shown. These commonly used miniplates are straight titanium plates, consisting, depended on the type, of at least two finishing ring sections at both ends, where screws for an implant-bone fixation can be drilled through. The raw implant is available in different variations, consisting of further middle-ring sections, orthogonal bendings or circular ring section arrangements. Whereby, in this work, the focus lies on those which are commonly used in cranio-maxillofacial surgery, e.g. in cases of mid-facial or also partly in lower jaw trauma treatment.

**Table 1. Summary of results according to the trauma database due to facial injuries caused by ski accidents in Innsbruck (Department of Oral and Maxillofacial Surgery, Innsbruck, Austria).**

| Variable | 579 Skiers / 882 Incidents |
|---|---|
| **Age** | |
| Mean | 28.35 |
| Standard Deviation | 15.78 |
| Min | 2 |
| Median | 26 |
| Max | 81 |
| **Gender** | |
| Male | 378 |
| Female | 201 |
| **Injury Type** | |
| Soft tissue injuries | 336 |
| Facial bone fracture | 310 |
| Dentoalveolar trauma | 236 |
| **Mechanism of Injury** | |
| Collision with others | 135 |
| Struck by equipment | 70 |
| Falls | 263 |
| Collision with object | 46 |
| Lift accidents | 34 |
| Others | 31 |

https://doi.org/10.1371/journal.pone.0182839.t001

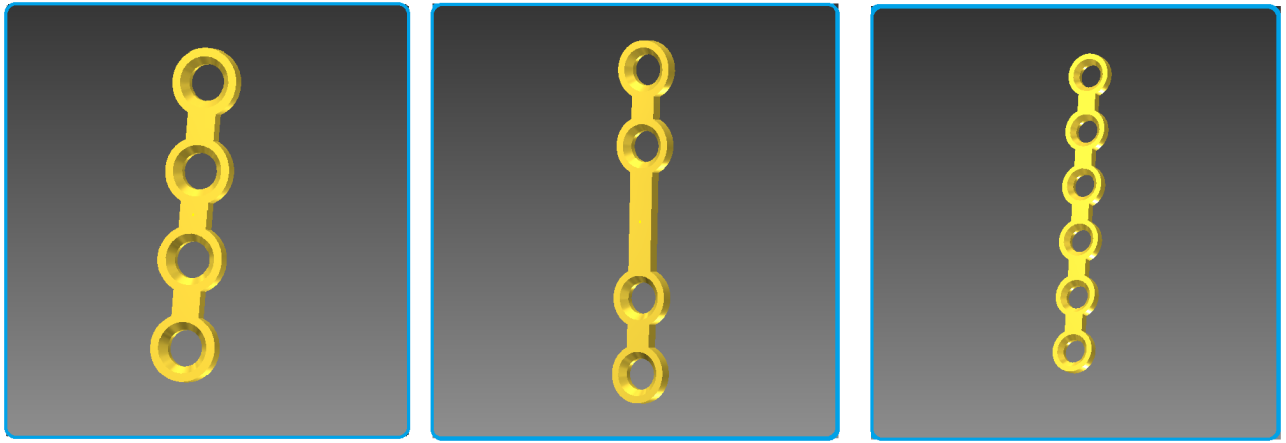

**Fig 1. The three figures show commonly available miniplates for bone fracture fixation.** All models are of the MedArtis Trauma 2.0 series. The most left one, with article number M-4138, shows an overall length of 23 mm consisting of two end ring elements and two middle ring elements. The model in the middle, with article number M-4320, with the same properties as the left one but an extended bar of 9 mm and thus extended to 29 mm in the overall length. Last, the most right one, with article number M-4322, with two additional ring elements and an overall length of 35 mm.

The Modus 2.0 series from the MedArtis Group (www.medartis.com) were chosen as the osteosynthesis materials used in this trial, (Fig 1).

For facial reconstructions, the implants are fixed perpendicular to the fracture on both fracture sites with the purpose of stabilizing the defect. In general, the miniplates cannot be deployed in their stiff and straight initial form why it is necessary to adapt them along the fractured surface to gain a perfect fit using a bending tool. This can lead to intraoperatively performed time-consuming bone plate adapting procedures, especially in complex surgical cases. Another method is to generate a stereolithographic (STL) model out of the Computed Tomography (CT) scan data and pre-bend the implant on those [6]. This method takes even more time, especially in the pre-operative phase, also leading to higher expenses in overall costs [7]. Further, there are commercial software tools available, like offered by Materialise (www. materialise.com). However, these software packages include complex user interfaces that are difficult to handle in the practical routine and require high monetary costs since they are not licensed free, personal resources for their use or further outsourced services to the industry. Further, with this software it is not possible to pre-plan the application of commercial available osteosynthesis materials, since the software is created to generate a patient specific implant that is offered and manufactured by the industry as a high monetary service and sold to the clinical center. Therefore, this contribution proposes a new method for computer aided planning of facial surgeries with miniplates using the medical image processing platform MeVisLab (www.mevislab.de), which we have already successfully used in various medical applications, like [8–13], and integrates and builds upon the common ITK and VTK libraries, similar to (3D) Slicer [14], [15]. By using this software, the surgeon has the possibility to plan the implant independent of its position on the facial bone without using a 3D model. Further, the planning time can be performed in only a few minutes and readjustments of the material position are easily possible to perform without any additional costs. This is achieved by first calculating the surface curvature of an initially set marker on which the further implant is created. The generation includes the alignment and orientation of pre-constructed osteosynthesis models, for the ring sections of the bone plate, and the generation of the bridge sections using Delaunay-triangulation [16]. Finally, the implant model can be 3D-printed [17–19], since it is stored in STL-file format, usable as bending and adapting tool or even as the final created implant.

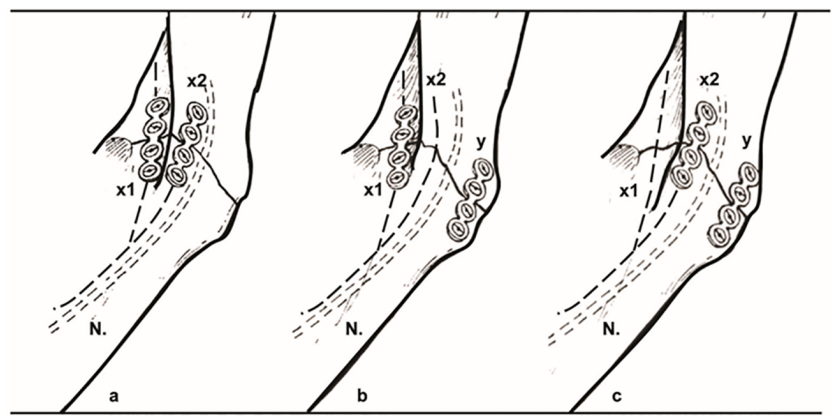

**Fig 2. a,b,c: Clinically used osteosynthesis of the left mandibular angle using two plate techniques.**
X1, x2 plate position, N: Inferior alveolar nerve in the mandibular bone. Note: The nerve does not interfere with the bone plates to avoid nerve damaging and loss of lip sensibility.

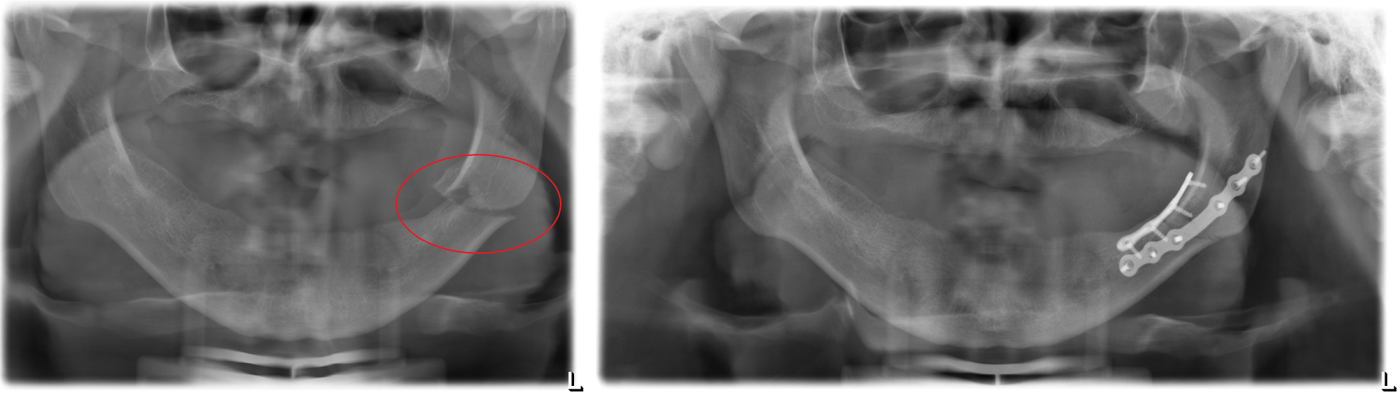

**Fig 3. Image of real patient x-ray data suffering from a mandibular angle fracture (left angle).** Left image shows the untreated defect. The result of the treatment with applied miniplates after bone repositioning is shown on the right part of the figure.

Others working in the area of computer-aided treatment of facial bone defects are Zhao et al. [20] who present the application of virtual surgical planning (VSP) with computer-assisted design (CAD) and manufacturing (CAM) technology to cranio-maxillofacial surgery. In doing so, they illustrate the components, system and clinical management of the VSP and CAD/CAM technology, which includes also the data acquisition, virtual surgical and treatment planning, and fabrication, and outcome assessment. However, they focus more on individual implant design, instead of common available implants. Further, Parthasarathy [21] gives a review about recent trends in 3D modeling and custom implants in craniofacial reconstruction. Bell [22] presents and overview of computer planning and intraoperative navigation in cranio-maxillofacial surgery. The author states also that in situ plate bending is time consuming, and that in situ plate bending is not practical if tumor grossly invades soft tissue on the lateral mandible. Moreover, the use of stereolithographic models for plate adaptation before surgery has been used by the author for the past 5 years to aid in maxillo-mandibular reconstruction. Furthermore, Adamus and Lacki [23] investigated the titanium bending process. Therefore, they carried out a numerical simulation based on the finite element method of the bending of a Ti6Al4V ELI titanium alloy bar with the Adina System. Summarized, the influence of bar diameter, bending radius and bending angle on the strain and stress distribution in the deformed element was analyzed. However, they state that problems occur when bent titanium elements have to be bent out or bent over during their exploitation, e.g. some titanium body implants (like miniplates) need to be manually adjusted to the bone curvature during surgical operations. The work closest to ours is from Burghart et al. [24] who present a planning system for the preoperative positioning of miniplates on a model of the patient's skull. However, they don't report how fast the miniplate model calculation is and if it enables and interactive placement in all steps. Furthermore, their tool is not available for other researchers and it is not clear how many (medical) datasets they used for testing, and if they had typical cases from the clinical routine. Finally, it seems they did not model real and conventional available miniplates from a manufacture via their blueprints as it is performed in this study. All this shortcomings will be addressed in this contribution and thus it is a consequent further development of their research work.

This contribution is further organized as follows: Section 2 depicts details of the used material and the newly proposed integration, presents the theoretical background of the proposed mechanism and provides sufficient detail to allow the work to be reproduced. In Section 3, experimental results, including illustrations of generated implants, are presented. Section 4 gives a summary together with a short discussion to highlight the significance of the introduced achievements and lays the foundation for further work.

## Material and methods

### Datasets

Only high-resolution data sets (512x512 voxel in x- and y-directions) with slices not exceeding 1.0 mm with 0.25 mm pixel size in z-direction and providing physiological, complete mandibular bone structures without teeth were included in the selection process. Further, no difference was made between atrophic and non-atrophic mandibular bones—both were included during the selection process. However, incomplete data sets consisting of mandibular structures altered by iatrogenic or pathological factors or fractured mandibles as well as data sets showing osteosynthesis materials in the lower jaw were excluded in this trial. The used data sets were provided as DICOM-image files [25]. The proposed software, however, uses the STL file format why a conversion was performed using a simple MeVisLab network. This operation also holds the positive side effect of patient information stored in DICOM tags is lost leading to anonymized datasets. In addition, teeth and alveolar processes were removed due to image-based artefacts to create an objective collection of data samples. For a further assessment or own research purposes some of the data used in this trial can freely be downloaded, but we kindly asked to cite our work [26]: https://figshare.com/articles/Cranial_Defect_Datasets/4659565

Note: when more datasets get available over time or other researchers provide us new cases, we will add them to the collection.

Ethics statement: All patient data was anonymized prior to analysis by the authors. In addition, the physicians and doctors who tested our newly developed tool have consented to their results of testing being published.

### Miniplate models

Purchasable miniplates from the Modus 2.0 series from the MedArtis Group (www.medartis.com) were chosen as the osteosynthesis materials used in this trial [27]. Therefore, we created CAD models reverse engineered from the miniplates blueprints. Three CAD models for common miniplates for bone fracture fixation from the Trauma 2.0 series are shown in Fig 1. The most left one, with article number M-4138, shows an overall length of 23 mm consisting of two end ring elements and two middle ring elements. The model in the middle, with article number M-4320, with the same properties as the left one but an extended bar of 9 mm and thus extended to 29 mm in the overall length. Last, the most right one, with article number M-4322, with two additional ring elements and an overall length of 35 mm.

## Workflow overview

In this contribution, we implemented a MeVisLab network that generates and applies osteosynthesis materials such as commonly used miniplates. The materials, based on a seed point, can be placed by the user, in a two-step approach. First, the user has to load the dataset into the application to be able to set an initial point defining the implant's center position, followed by setting up the direction and choosing the appropriate implant model necessary for calculating a baseline. This baseline shows the implant's centerline [28], [29] and serves as a simplified model saving computational time but still giving the user a clear and accurate response. If satisfaction is reached, the user visualizes the implant in a second step, which is further able to be saved for parallel implant generation, or exported, for local and permanent storage. In the next section this workflow is described in more detail, explaining the network and its single modules.

## Network

Fig 4 shows the constructed network, created within the MeVisLab platform. MeVisLab provides basic and advanced algorithms for medical image processing and visualization and also includes an environment for individual module generation under C++ as well as an environment for implementing own user interfaces with the MDL-script (MeVisLab Description Language). Each block of the modular pipeline network represents a module with individual functions. Three different colors represent the different types of available modules, where blue represents the procession of WEM data, green modules mark an Open Inventor module, working with visual scene graphs (3D), and last, the brown modules named macro modules

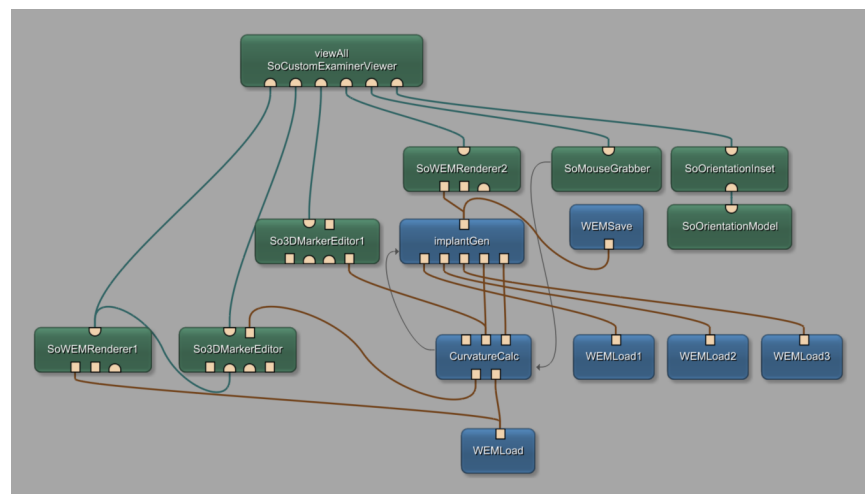

**Fig 4. Constructed network using the MeVisLab prototyping environment, including self-implemented modules named *CurvatureCalc* for baseline calculation and *ImplantGen* for generating the implant models.**

https://doi.org/10.1371/journal.pone.0182839.g004

hiding sub networks for preserving a clear network structure. Connections between modules use two different connector types: Undirected lines, for a data connection, and directed ones, indicated by arrows, for a parameter connection. Using data connections, it is important to connect the output of one module, with an input of the same type of another module (note: the data flow is bottom-up). Also, in- and outputs are distinguishable by three different types: The squared ones for Base-objects, pointing to data structures, the triangulated ones for ML images, and half circled ones for Inventor scenes. With the introduced modules and connections, the network was set up as follows: Starting at the bottom of the network, the *WEMLoad* module loads the dataset into the network which is immediately rendered with the *SoWEM-Render1* module and visualized in the main viewer panel by the *viewAll SoCustomExamin-Viewer*. The self-implemented *CurvatureCalc* module holds the implementation for generating the baseline. Therefore, in addition to the dataset itself, the network uses the user set point, marking the implants center. This point is obtained by the *So3DMarkerEditor*, handling user interactions in the viewer panel. The two outputs of the *CurvatureCalc* module hold once the 3D-position values for the baseline on the left, and once the according normal vectors at each point on the right. Since the baseline is visualized continuously we use another *So3DMarkerEditor* named *So3DMarkerEditor1*. Next, the self-implemented *ImplantGen* module derives the three-dimensional position of each baseline point and the according normals for generating the final implant. Further, necessary do accomplish this task, the pre-constructed models of the ring section, loaded with the additional *WEMLoad* modules, are connected to the input. The additional *SoWEMRenderer2* renders the generated implants which are passed on and visualized together with the dataset. With the *WEMSave* module, it is possible to store the generated miniplates locally in STL format for further use or for 3D printing. Yet, three modules are not described. *SoMouseGrabber*, *SoOrientationInset* and *SoOrientationModel* handle the rotation of the mouse wheel which's value is passed to the *CurvatureCalc* module using a parameter connection. In addition, the two self-implemented modules are connected with a parameter connection, which is important to further pass on the selected implant type.

## User interface (UI)

As unintended user interaction with the network may lead to network errors, we packed the constructed network into a single macro module. The module also holds a MDL script file implementing the user interface (UI) as illustrated in Fig 5. The UI is separated in two main parts: The control panel, including all interaction fields which actual change the data, and the viewer panel including all visualizing interactions.

## Generating pipeline

Fig 6 illustrates the workflow pipeline which can be separated in three groups according to the two procedural steps, implant setup and implant generation, and the initial data acquisition, all described in detail in the following paragraphs:

**1. Data acquisition—Orange.**  First, the user has to load the patient's dataset by choosing a path to the locally stored WEM file by clicking on the browse button of the *Loadfile* field in the user interface, resulting in already visualizing the data set in the user interface's viewer panel. The viewer implements already some camera controls, like zooming and rotating the camera.

**2. Implant set up—Grey.**  Second, the user sets up the implant's properties by defining the center point of the implant through placing an initial point on the skull. This is accomplished by holding the ALT key and clicking on the desired location on the 3D models surface. The Baseline Calculation algorithm (described in section *Module: CurvatureCalc*) starts now

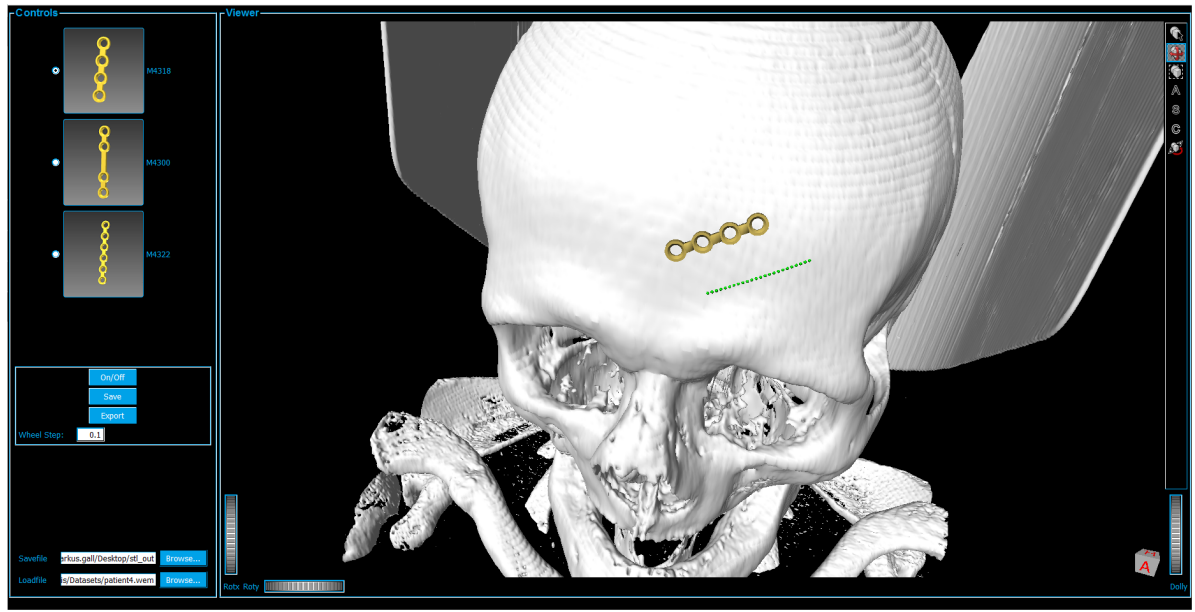

**Fig 5. User interface, separated in controls and viewer panels.** The Viewer box shows the data set (white skull) attached with a miniplate (gold) and the baseline (green).

https://doi.org/10.1371/journal.pone.0182839.g005

calculating the Baseline (green dotted line) with an initial orientation that depends on the mouse wheel value and the position of the central point. This baseline shows the location, direction and the approximated length of the later generated implant including the underlying curvature and serves as a simplified version of the resulting implant. Hence, giving already an idea on how the implant will be applied. In Fig 5, the resulting baseline (green dotted line) is illustrated where the center position is placed on the forehead. Further, the length of this line depends on the selection of the implant model, by marking one of the radio buttons in the user interface's control box on the left side. Since they vary in length, different models set the baseline's length to different values according to the model's dimensions. Additionally, by turning the mouse wheel, the implant can be rotated, thus the direction of the baseline alters to match the users desired placement and orientation. The field *wheel step* lets the user choose the accuracy of this rotation for high precision positioning. Also, the initial point, and thus the implant itself, can be readjusted at any time by holding the ALT key and clicking on the new position on the data set's surface. We chose the intermediate step of visualizing the baseline first rather than calculating the full implant directly to gain a more comfortable user experience since the generation of the model utilizes several seconds of time resources, resulting in unwanted latencies that we want to avoid. With this compromise, the user gains a fast visual response of the implant's placement and adjustments are realized close to real-time. Moreover, this method does not lack any characteristics a fully generated implant model would provide, as the baseline holds all the necessary information for optimal placement of a miniplate. If satisfaction concerning the setup is reached, the generation of the implant follows, as described in the subsequent section.

**3. Implant generation—Apricot.** This step uses the baseline that is setup in the previous section to generate the final implant model. Actuating the *ON/OFF* toggle button in the UI initializes this operation. In the overview workflow diagram this interaction accords to the green, diamond shaped block named *Visualize ON/OFF*. In addition to the baseline properties position, direction and implant model, this algorithm uses the implant's ring sections, namely the end and middle elements for constructing the final implant (details can be found in section *Module*: *ImplantGen*). Depending on the implant model's dimensions the positions and orientations of each ring element are sequenced calculated followed by further generating of the bridge sections, also according to the model's specifications. The ring elements have been pre-constructed in Autodesk's Inventor (http://www.autodesk.com/products/inventor/overview). Again, Fig 5 shows a final result of this procedural step where the implant is visualized as a golden structure. Currently, for the proof of concept, the selection of available implants is limited to different models from the Medartis Modus 2.0 series, which are characterized in the product catalogue. Despite the fact that the implant's setup can be altered during every step, it has to be kept in mind, that rotating the implant or adjusting the position in visualizing mode takes a lot of computational afford and thus may result in latencies. For optimal usage, it is therefore recommended to adjust these characteristics only during inactivated visualization. In the case of observing a satisfying result, the user can save the implant to continue setting up another implant while visualizing the previous generated and saved ones. This is accomplished by using the *Save* button, resulting in saving the implant for the current session only. This method allows the generation and visualization of multiple implants, or in other words, this function allows visualizing more than one implant at a time and is important in medical cases where fracture treatment uses more than one miniplate as it is the case e.g. by suffering from LeFort fractures [30]. However, it is important to understand the difference between the *Save* and *Export* buttons. The *Export* button does export all implants, including the current (may not be saved yet) and all saved ones to the path, given in the *Savefile* field, as STL file. Again, export actually also means to save, but saves the implants permanently on the local drive in STL format, rather than the *Save* button which does save the currently processed implant for this session to the array of other already saved implants. This means, that as long as an implant is not saved, it is possible to be readjusted. Another option, which is permanently able to be used, as long as the baseline is already generated, is to change the baseline's marker points in any arbitrary direction by just changing its position using drag and drop interactions.

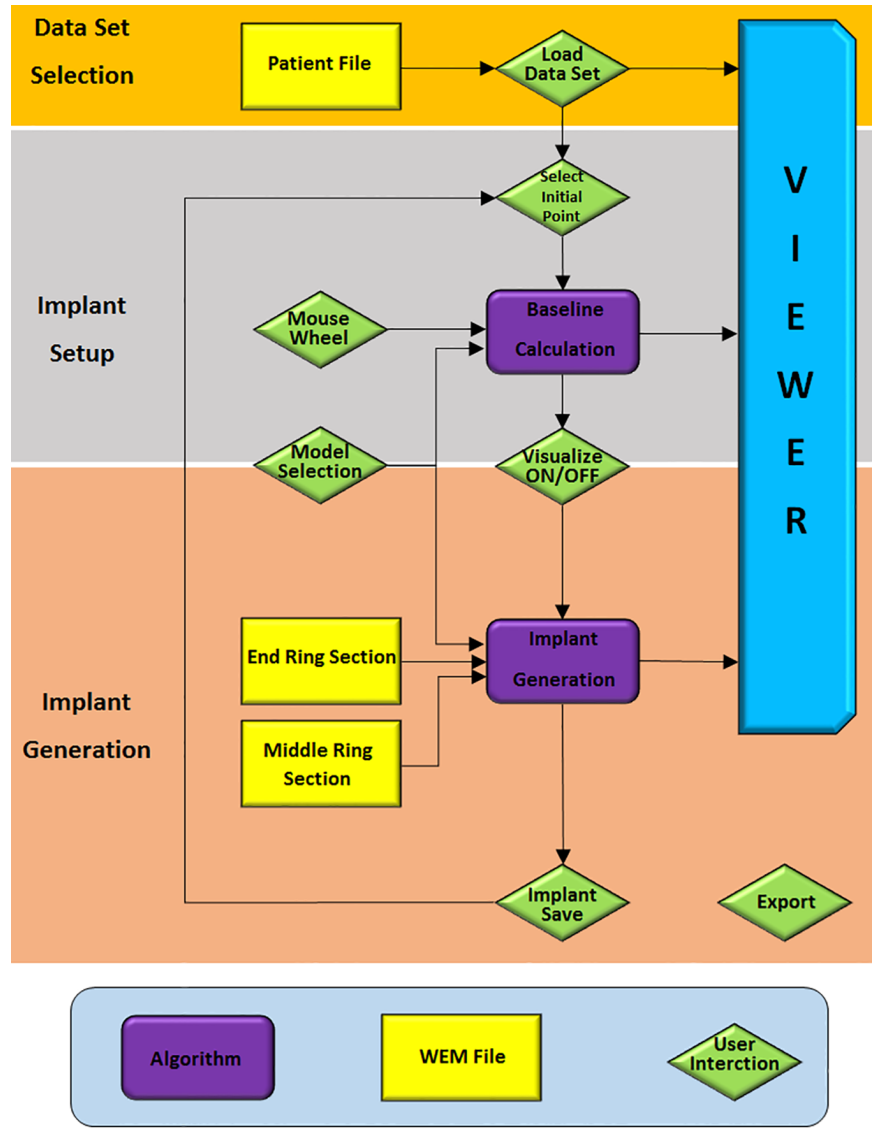

**Fig 6. Flow chart of a miniplate generation process separated in three main parts: Orange marks the data acquisition, grey the implant set-up including the baseline calculation and apricot the implant generation and visualization.**

Within this section the reader should have gained an understanding on how the software tool works and is used for planning the adaption of osteosynthesis material in the face. However, together with the description of the self-implemented algorithms of the *CurvatureCalc* and the *ImplantGen* modules in the following sections, the reader gets a comprehensive description on how the underlying processes are assembled.

**Module: *CurvatureCalc*—The baseline that is visualized as a very basic structure in form of markers is calculated in the *CurvatureCalc* module. The baseline is a simplified model of the final implant but already showing the most important characteristics like position, direction and curvature. Since the calculation of the baseline is less consuming in computational costs, the generation and adjustment of the baseline happens much faster than a direct visualization of the final implant. The method is based on the work of Möller and Trumbore [31], [32] and their calculation method of ray-triangle intersections. The method casts a ray from the origin $O$ of the cast ray in direction $D$ and then changes the base of that vector that yields a vector $(t\ u\ v)^{\mathrm{T}}$, where $t$ represents the distance to the plane in which a triangle lies and $u,v$ as the coordinates inside the triangle. The big advantage is that the computation of the plane equation is not performed resulting in significant memory savings. This method is used to determine the baseline where not only one ray but also a cascade of parallel rays is cast, like shown in Fig 7. The object's surface, or in other words the patient's skull, is represented by the light blue triangulated mesh. The red star in the figure illustrates the initially set point as a starting value. From this point, first the normal vector $N$ (purple) according to the nearest triangle is elicited. Next, the direction vector $D1$ is determined by calculating the cross product of the normal vector $N$ (which is the same for the implant and the baseline position) and the baselines direction vector $D$, which are not necessarily perpendicular to each other. The origin of $D2$ is then calculated according to the implant model's measures, parallel to $D1$. Starting from this vector's origin, a cascade of rays is cast along vector $D2$'s direction, which means that the origins of the rays, used for ray-triangle intersection, are located along vector $D2$. The next step is to cast rays from the mentioned origins in direction of the negative normal vector $-N$. Now, according to Möller and Trumbore, the ray-triangle intersections can be located to set marker at those points (light blue stars), representing the final baseline. As a result, the baseline points' positions in the three-dimensional space are applied to one output of the module as a vector where each entry holds the three-dimensional vector to the points' position. Additionally, the according normal vectors to each baseline point are connected to the second output, again as a vector where each entry holds a three-dimensional vector.

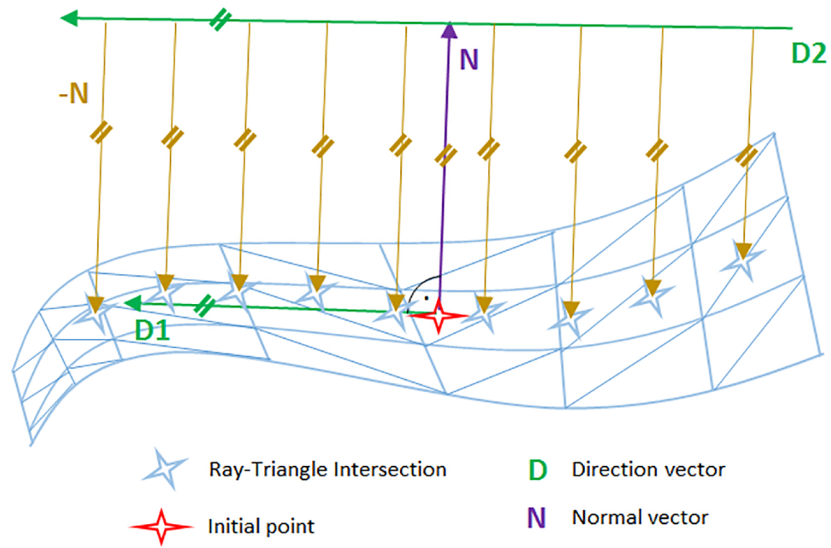

**Fig 7. Illustration of the baseline determination.** The direction vector *D* (green) is set up parallel to the initial point's (red star) direction vector (green). The direction vector is perpendicular to the initial point's normal vector (purple). The origins of the cast rays (brown) are translated along the direction vector and checked for intersections.

Module: *ImplantGen*–This module uses the information derived from the *CurvatureCulc* module described in the previous section to set up the final miniplate model in a piecewise manner. Therefore, first two ring sections are brought into position followed by building the connecting bridge section rather than building the first bridge element after the first ring element. The generation process is executed not only with respect to the baseline characteristics, but also includes the information of the basepoint's normal vectors, the selected implant type and the pre-constructed ring section models. The specific implant model distance from the center (initial marker position) along the baseline is calculated for the exact placement of the first ring element. Then, this point serves as a starting point where the first ring is located. Following ring sections are placed again along the baseline but in opposing direction where the previous located element is used as starting point for distance calculation. Correct alignment of orientation and direction of each section is ensured by using the normal vector marker list, where the ring element's normal vector *N* is adjusted to align the baseline's normal vector *N* at the position of correct placement, thus ensuring perfect alignment. Since the ring sections are not symmetric due to at least one plane side for bridge section attachment, this "open" sides need to be orientated correctly as well. Therefore, the direction vectors *D* of both, the ring element and the current baseline position are utilized where the direction vector of the baseline shows the direction to the next marker and the implant's direction vector describes the connecting end of the ring. Fig 8 illustrates this idea by showing the baseline (green) on the patient's skull, together with a ring element and described vectors *N* and *D*. For correct orientation of the ring elements, the two direction vectors are aligned in only two dimensions. Important to note is, that this is not a crucial task, since the baseline's normal vector and direction vector are not perpendicular to each other, which is different to the ring section where this is always the case. Therefore, the direction vectors are only aligned in two dimensions since it is important to keep the normal vectors aligned in all three dimensions. A final alignment shows all four vectors in the same plane which is spanned by the two normal vectors, congruent with each other, the perpendicular direction vector of the ring model and the baseline's direction vector. This is how each ring element is aligned and orientated correctly depending on the underlying surface geometries. However, to finish an implant, the bridge sections between those rings elements need to be generated as well: For building these connections, we use at least two triangulated blocks per sections. These blocks are connected to the corners of the plane sides of each ring element, illustrated as green crosses in Fig 9. Thus, the diameter shows a rectangular shape. First, the four corner positions of where the bridge elements are connected to the ring elements are determined. Since the ring's geometries are known, it is sufficient to forward the rings center position as well as the normal and the direction vector each, to the determining method, which applies distance measurements using the L2-norm. Next, at each baseline point between those two plane sides (green dots in Fig 9) a rectangle with the same dimensions as the ring element's plane rectangle is spanned (light blue rectangles in Fig 9). The orientation is defined by the basepoint's direction and normal vector. As a result, the basis for finalizing the bridge element using triangulation is built. In a next step, the starting, ending and the corner positions of the rectangles are used for the triangulation task. To do so, first, a surface consisting of four nodes has to be defined. As an example, one uses two neighboring nodes from the starting rectangle (*N1* and *N2*) and the opposing ones of the first constructed rectangle (*N3* and *N4*) where further triangulation is performed. This operation is performed for all four surfaces of each block resulting in the final implant.

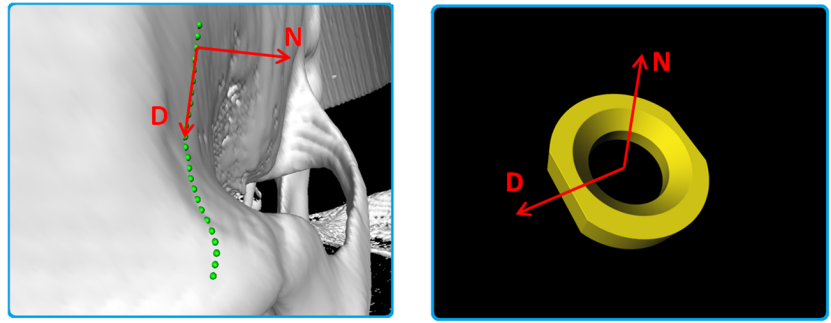

**Fig 8. The patient's skull (white) with an applied baseline (green) are shown on the left side, where the normal vector _N_ and the direction vector _D_ are drawn for the placement of the ring elements.** The ring element itself is shown in the right image, where the normal vector _N_ and direction vector _D_ (red), which have to be aligned with the vectors from the baseline, are illustrated.

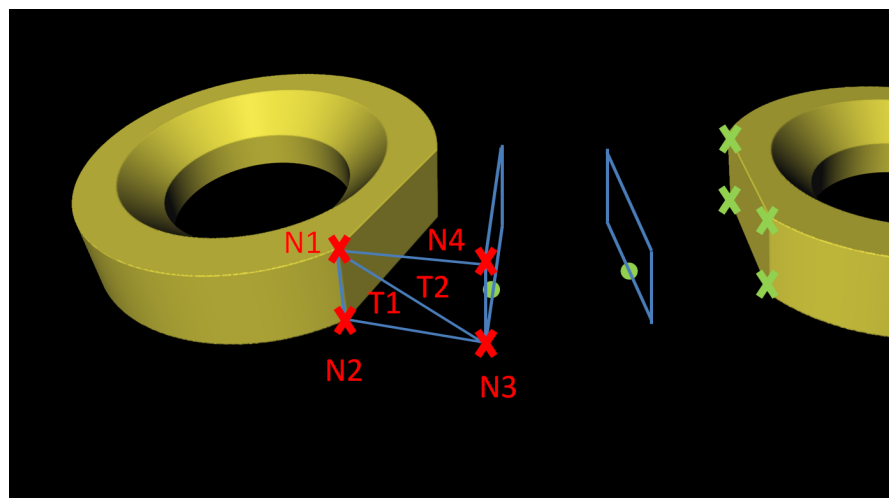

**Fig 9. Bridge building principle with triangulation.** Green crosses mark the corner of the ring element's plane side. To each baseline point (green points) a rectangle (light blue lines) is constructed. Rectangle triangulation, spanned by the two starting nodes (_N1_ and _N2_) and two opposing nodes of the previous constructed rectangle (_N3_ and _N4_). Node _N1_, _N3_ and _N4_ build a triangle _T2_, and Nodes _N1_, _N2_ and _N3_ build the second triangle _T1_.

## Results

The presented software was evaluated using ten CT data sets performed out of diagnosis and treatment reasons in the clinical routine at a department of Oral and Maxillofacial Surgery. The study participants for our software were clinical experts from the Department of Oral and Maxillofacial Surgery, Medical University of Graz, Austria and technical experts from the Institute of computer Graphics and Vision from the Faculty of Computer Science and Biomedical Engineering of Graz, University of Technology, Austria (TU Graz). We included physicians and computer scientists into the study to have different points of view on the software solution and its interface. However, all experts are familiar with the state of the art techniques in facial reconstructions and the performed evaluation consisted of three main steps. First, during the so-called training phase, the developer of the prototype explained the software by demonstrating an example planning case and explaining the functionality of the software. In the next step, the physicians performed the planning of an implant on their own, where we described the case to be solved with the following words:

> Load the given data set and perform the placement of a mandibular media fracture using at least two implants of different model type and export the data to the desktop, if satisfaction is reached.

Subsequently, the testers were asked to answer a questionnaire containing eleven questions on a six point Likert-scale with increasing consent from one to six [33] to rate the overall software solution. Generally, the questionnaire evaluates the software ergonomic formulated in general terms without reference to situations of use, application, environment or technology [34]. However, questions like "The software does not need a lot of training time" are seen in context of the current user and his/her experience/knowledge in miniplate planning (in respect to other tools or their medical experiences). One question (question 11) had to be answered with *yes* or *no*. The questions are listed below and are adopted from the software evaluation questionnaire of ISO Norm 9241/110 [35].

1. The software does not need a lot of training time

2. The software is adjusted well to achieve a satisfying result

3. The software provides all necessary functions to achieve the goal

4. The software is not complicated to use

5. How satisfied are you with the UI surface?

6. How accurate was the placement of the implant?

7. How satisfied are you with the presented result?

8. Was it easy to adjust the implant? (position, orientation, model, ...)

9. How satisfied have you been with the time consumed? (no training)

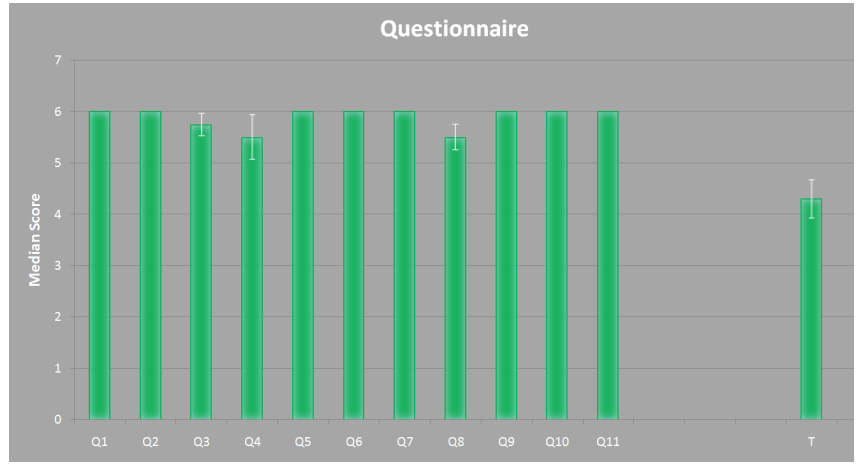

**Fig 10. Result of the questionnaire derived from Table 2.** The *x*-axis represents the questions and the time. The *y*-axis shows the rating score of each question together with the superimposed standard error.

10. How is your Overall impression

11. Would you use the software in a daily routine?
    (Assumed that the 3D printed implant would fit with just a few adjustments)

12. Time used for gaining a satisfying result

The result of this questionnaire is shown in Fig 10. The *x*-axes represents the questions from one to eleven (Q1—Q11) and the measured time *T*. The *y*-axis represents the median score for Q1 to Q10. In addition, the median value for Q11 is shown but represents a *yes* or *no* answer, where *yes* equals a value of six points and *no* equals a value of zero points. The time is represented in minutes. For all data the superimposed standard error is represented by the white lines. In Table 2 the underlying data is represented, where subject *two* and *three* represent the surgeons' results.

**Table 2. Results from the evaluation questionnaire for each subject using a six point Likert rating.**
*Median* is the median value of the four subjects and *Error* describes the superimposed standard error. *Q11* is a *yes/no* question where *no* equals a value of *zero* and *yes* equals a value of *6*. The time *T* is measured in minutes.

| No. | Subject | | | | Median | Error |
|-----|---|---|---|---|--------|-------|
| | 1 | 2 | 3 | 4 | | |
| **Q1:** | 6 | 6 | 6 | 6 | 6.00 | 0 |
| **Q2:** | 6 | 6 | 6 | 6 | 6.00 | 0 |
| **Q3:** | 6 | 5 | 6 | 6 | 5.75 | 0.21 |
| **Q4:** | 4 | 6 | 6 | 6 | 5.50 | 0.43 |
| **Q5:** | 6 | 6 | 6 | 6 | 6.00 | 0 |
| **Q6:** | 6 | 6 | 6 | 6 | 6.00 | 0 |
| **Q7:** | 6 | 6 | 6 | 6 | 6.00 | 0 |
| **Q8:** | 5 | 6 | 6 | 5 | 5.50 | 0.25 |
| **Q9:** | 6 | 6 | 6 | 6 | 6.00 | 0 |
| **Q10:** | 6 | 6 | 6 | 6 | 6.00 | 0 |
| **Q11:** | 6 | 6 | 6 | 6 | 6.00 | 0 |
| **T:** | 4 | 5.5 | 3.5 | 4.2 | 4.00 | 4.30 |

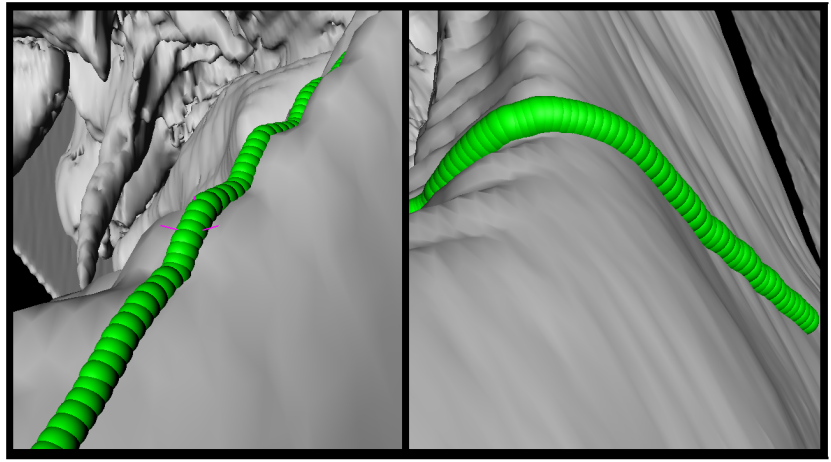

**Fig 11. Baseline (green) on more complex surfaces.** On the left image a wave like surface is shown, and on the right side the baseline adjusts to a ninety degree angle.

Moreover, the baseline determination is very robust and is designed to capture the curvature of the skull's surface independent of the surface curvature. Fig 11 shows that the baseline does adjust itself also on very difficult surfaces like a wave form or even 90 degree angles. In Fig 12, an example result is shown, containing three available implants, each located on a position where they are generally applied (left). For a better illustration, the single implants are extended portrayed on the right side as well. Additionally, Fig 13 shows the same outcome from its sagittal and coronal perspective.

## Discussion

In the proposed work, a workflow and software tool for position planning of miniplates to treat facial bone defects and reconstructions was implemented, which can be used in the clinical routine. Therefore, a modular network with the medical imaging platform MeVisLab was constructed that is able to load a patient's CT-data set and plan the positioning of commonly used miniplates.

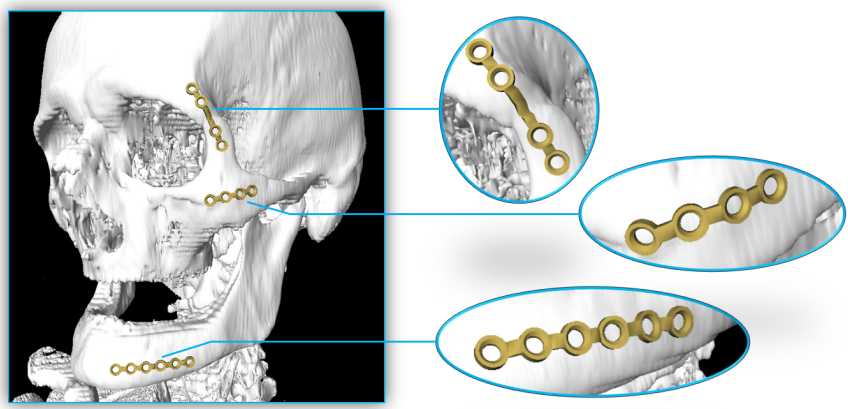

**Fig 12.** Example result containing all three available implants each located at a position where they generally are applied (left). For better illustration, the single implants are extended portrayed as well on the right side. The plate could easily be positioned according to the underlying bone contours by the used software module.

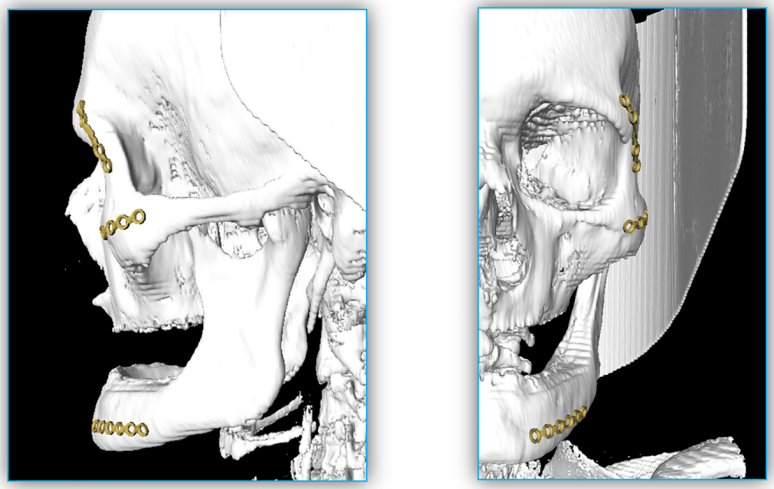

**Fig 13. Result from Fig 12 in its sagittal (left) and coronal (right) perspectives.**

The application of miniplates in the facial area can be challenging especially in complex surgical cases. This lacks not only on appropriate (free) software, but also state of the art techniques show some space for improvement. Furthermore, commercial software products do not provide an option for the application of commonly used osteosynthesis material but rather focus on the generation of patient specific implants manufactured and sold by the industry. In this contribution, we have shown that with the use of MeVisLab it is possible to generate a custom network allowing preoperatively an easy and precise virtual placement of commonly used miniplates. Furthermore, it is possible to use the outcome together with 3D printing technologies since the planned implant can be exported as an STL-file [17]. The 3D printed implant can be used for an intraoperatively performed direct adaption of the bone plate according to its function as a bending tool. In summary, the proposed contribution achieved the following research highlights:

- The successful interactive planning and reconstruction of facial 3D implants;

- A precise capture of the surface curvature as a basis for any type of implant;

- Flexibility in model types as well as in positioning and orientating the miniplates;

- Evaluation with real patient Computed Tomography (CT) data from the clinical routine;

- Enabling 3D implant export as Computer-Aided Design (CAD) file format for 3D printing and in-depth pre-surgical assessment;

- Providing clinical datasets and source code to the research community for own usage.

The intention of this contribution was to work out a workflow and establish, and evaluate a software for the position planning of miniplates. A next step will be to test the software in the clinical routine by using the planned miniplate positions in operations. Future works include also the implementation of further implant types and even patient specific ones [36] to provide to the user an even higher flexibility. In addition, bone repositioning [37] is of high priority in future research, since fractures in the facial area mostly suffer from dispositions where miniplates may not be deployed without further ado. Moreover, performing the miniplate generation on the GPU [38], [39] to enable an interactive placement of the implant and thus skipping the baseline step. Finally, it is necessary to perform a more comprehensive evaluation of the software/workflow and the planning could be performed in Virtual Reality (VR) [40] whereas the intraoperative deployment of the miniplates could be performed in an Augmented Reality (AR) setting [41].

## Acknowledgments

Blueprints of the miniplate dimensions were provided by Jan Stanzel from the MedArtis AG, Hochbergerstraße 60E, 5057 Basel, Switzerland. A tutorial video demonstrating the interactive planning of miniplates can be found under the YouTube channel: https://www.youtube.com/c/JanEgger/videos

## Author Contributions

**Conceptualization:** Jan Egger, Jürgen Wallner, Markus Gall, Knut Reinbacher.

**Data curation:** Jürgen Wallner, Xiaojun Chen, Knut Reinbacher.

**Formal analysis:** Jan Egger, Jürgen Wallner, Markus Gall, Xiaojun Chen, Knut Reinbacher.

**Funding acquisition:** Katja Schwenzer-Zimmerer, Dieter Schmalstieg.

**Methodology:** Markus Gall.

**Project administration:** Jan Egger, Xiaojun Chen, Katja Schwenzer-Zimmerer, Dieter Schmalstieg.

**Resources:** Jan Egger, Jürgen Wallner, Xiaojun Chen, Katja Schwenzer-Zimmerer, Dieter Schmalstieg.

**Software:** Markus Gall.

**Supervision:** Jan Egger, Xiaojun Chen, Katja Schwenzer-Zimmerer, Dieter Schmalstieg.

**Validation:** Jürgen Wallner, Markus Gall, Knut Reinbacher.

**Visualization:** Markus Gall.

**Writing – original draft:** Jan Egger, Markus Gall.

**Writing – review & editing:** Jan Egger.